\documentclass[runningheads]{llncs}
\usepackage{graphicx}
\usepackage{tabularx}
\usepackage{times}
\usepackage{caption}
\usepackage{subcaption}
\usepackage{epsfig}
\usepackage{graphicx}
\usepackage{amsmath}
\usepackage{amssymb}
\usepackage{array}
\usepackage{colortbl}
\usepackage{booktabs}
\usepackage{bbm}
\usepackage{multirow}
\usepackage{multicol}
\usepackage{makecell}
\usepackage{xcolor}
\usepackage{xspace}
\usepackage{float}
\usepackage{color}
\usepackage{enumitem}
\usepackage{pifont}
\usepackage[table]{xcolor}

\usepackage{epsfig}
\usepackage{graphicx}
\usepackage{amsmath}
\usepackage{amssymb}
\usepackage{booktabs}
\usepackage{tabulary}
\usepackage{dblfloatfix}
\usepackage{transparent}
\usepackage{algorithm}
\usepackage{listings}
\usepackage{algorithmic}
\usepackage{balance}

\usepackage{hyperref}
\usepackage[capitalise,noabbrev]{cleveref} 

\usepackage{listings}
\usepackage{bbding}

\lstdefinestyle{mystyle}{
    backgroundcolor=\color{gray!10},
    basicstyle=\ttfamily\small,
    frame=single,
    breaklines=true,
    columns=flexible,
    numbers=none,
    showstringspaces=false,
    tabsize=2
}
\AtBeginDocument{
  
}

\begin{document}
%
\title{ORMOT: A Dataset and Framework \\ for Omnidirectional Referring Multi-Object Tracking}
\titlerunning{ORMOT: Omnidirectional Referring Multi-Object Tracking}
%
\author{
Zihan Zhou*\inst{1} \and
Sijia Chen*\inst{1} \and
Yanqiu Yu\inst{1} \and
En Yu\inst{1} \and
Wenbing Tao\Envelope\inst{1}
}
\authorrunning{Z. Zhou et al.}

\institute{
National Key Laboratory of Science and Technology on Multi-spectral Information Processing,
School of Artificial Intelligence and Automation,
Huazhong University of Science and Technology, Wuhan 430074, China\\
* Equal contribution\\
\Envelope\ Corresponding author:
\email{wenbingtao@hust.edu.cn}}
\maketitle              

\begin{abstract}
Omnidirectional cameras provide 360° spatial coverage, making them increasingly valuable in applications such as autonomous driving and video surveillance. While Multi-Object Tracking (MOT) and its language-guided extension, Referring Multi-Object Tracking (RMOT), have achieved notable progress, existing methods rely on conventional cameras with limited fields of view, causing critical contextual cues to be lost when targets move outside the frame. This fundamentally limits the model's ability to interpret long-horizon language descriptions involving sequential actions, spatial relations, and group behaviors. In this work, we propose Omnidirectional Referring Multi-Object Tracking (ORMOT), a novel task extending RMOT to omnidirectional imagery, where 360° coverage ensures complete scene context for accurate language-guided tracking. To advance this task, we construct ORSet, a dataset comprising 27 omnidirectional scenes, 848 language descriptions, and 3,401 annotated objects. Furthermore, we propose ORTrack, an LVLM-driven framework that enables zero-shot language-guided detection and robust cross-frame association in complex 360° environments. Experiments on ORSet demonstrate that ORTrack achieves state-of-the-art performance, providing a strong baseline for future research. The dataset and code will be open-sourced at \url{https://github.com/chen-si-jia/ORMOT}.

\end{abstract}
%
%
%
\section{Introduction} 
\label{sec:Introduction}

\begin{figure*}[ht]
    \centering
    \includegraphics[width=1.0 \linewidth]{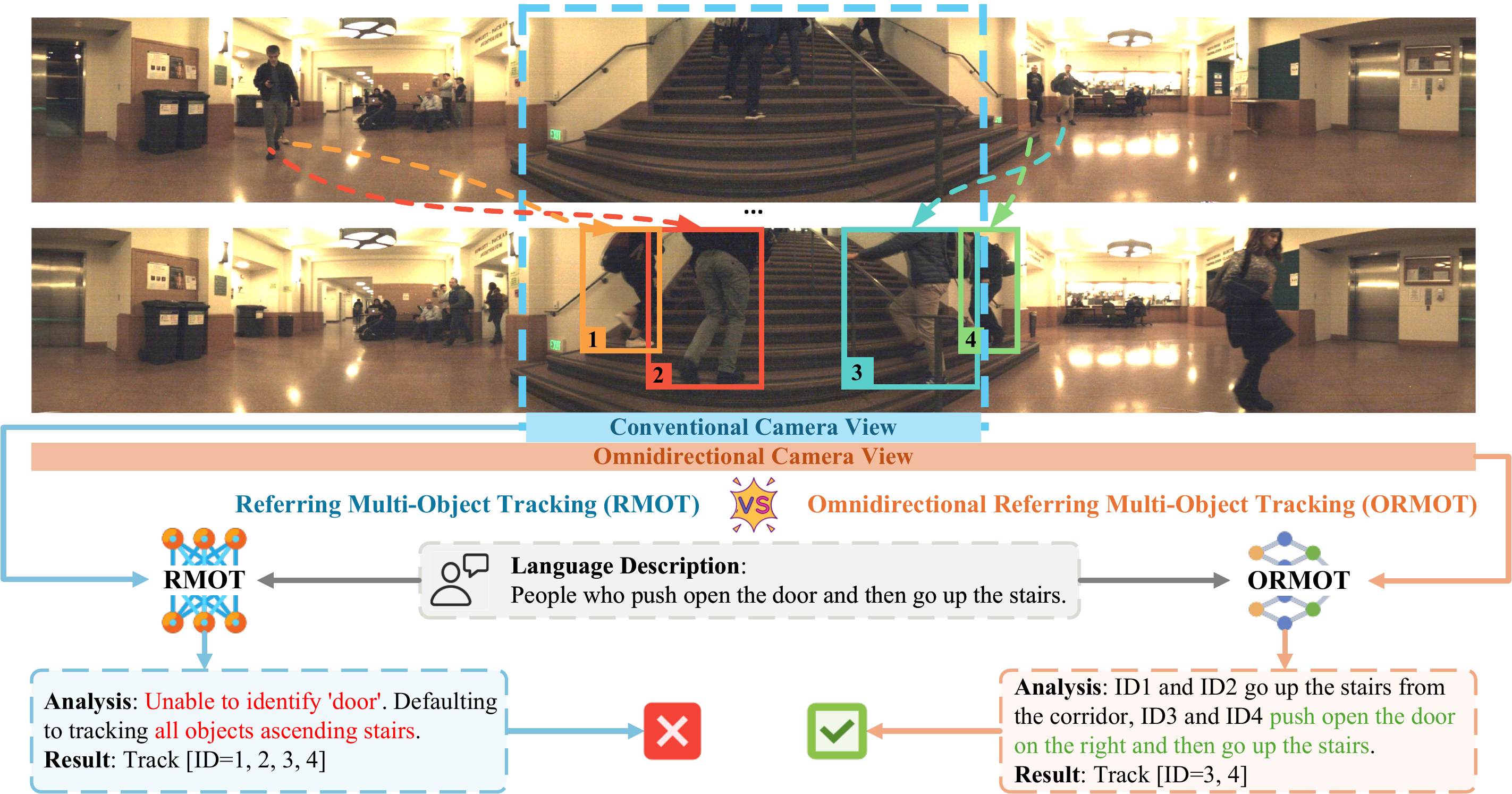}
    \caption{\textbf{Comparison between RMOT and ORMOT.} Given the description “People who push open the door and then go up the stairs,“ RMOT misses the door-pushing action due to limited field of view, incorrectly tracking all ascending people (ID=1,2,3,4). ORMOT captures the full scene and correctly identifies only the target objects (ID=3,4).}
    \label{fig:Comparison between RMOT and ORMOT.}
\end{figure*}

Omnidirectional cameras, with their 360° field of view, have become increasingly valuable in real-world applications such as autonomous driving \cite{meng2025motion} and video surveillance \cite{guan2025multi}, where complete spatial awareness is essential. In these settings, Multi-Object Tracking (MOT) serves as a fundamental perception task. To further align tracking with natural language, Referring Multi-Object Tracking (RMOT) \cite{wu2023referring} has been proposed, aiming to identify and continuously track objects that correspond to specific language descriptions.

However, existing RMOT methods are trained and evaluated on datasets captured by conventional cameras with a limited field of view. In real-world scenarios, language descriptions often involve long-horizon temporal and spatial contexts, such as sequential actions, spatial relations, or group behaviors spanning multiple locations. When key parts of a scene fall outside the camera's view, critical contextual cues are permanently lost, making it impossible to correctly interpret such descriptions. As illustrated in \cref{fig:Comparison between RMOT and ORMOT.}, given the description “People who push open the door and then go up the stairs," a conventional RMOT model fails to observe the door-pushing action when that area is out of view, and consequently misidentifies all people going upstairs as targets.

To overcome these limitations, we extend RMOT to omnidirectional imagery, establishing a new task termed \textbf{O}mnidirectional \textbf{R}eferring \textbf{M}ulti-\textbf{O}bject \textbf{T}racking (\textbf{ORMOT}). The 360° coverage ensures all relevant targets and contextual cues remain visible, while the broader scene captures inter-object relations and sequential actions critical for understanding long-horizon language descriptions. With these advantages, ORMOT models can correctly resolve descriptions like the above, accurately identifying the specific targets (ID=3,4) while avoiding semantic confusion.

To advance research on ORMOT, we construct \textbf{ORSet}, a dataset built upon JRDB \cite{martin2019jrdb} containing 848 language descriptions and 3,401 annotated objects across 27 omnidirectional scenes, providing a platform for research on temporal grounding and multimodal alignment. Furthermore, we propose \textbf{ORTrack}, an LVLM-driven framework that aligns arbitrary language descriptions with 360° visual scenes for zero-shot referring tracking, employing two-stage cropping-based feature extraction and cross-frame association via Hungarian matching \cite{kuhn1955hungarian}.

Finally, we evaluate ORTrack on ORSet under zero-shot conditions against representative open-source RMOT methods. ORTrack achieves state-of-the-art performance on both detection and association metrics, validating the effectiveness of the proposed framework for omnidirectional referring multi-object tracking.

In summary, our main contributions are as follows:
\begin{itemize} [leftmargin=6.3mm]
\item We propose \textbf{ORMOT}, a novel task extending RMOT to omnidirectional imagery, which improves the model's ability to understand long-horizon language descriptions by capturing complete spatial and temporal context.
\item We construct \textbf{ORSet}, a comprehensive dataset featuring 27 omnidirectional scenes, 848 language descriptions, and 3,401 annotated objects, serving as a benchmark for the ORMOT task.
\item We propose \textbf{ORTrack}, an LVLM-driven framework for ORMOT. Extensive experiments demonstrate state-of-the-art performance on ORSet, providing a strong baseline for future research.
\end{itemize}

\section{Related Work}
\label{sec:Related Work}

\subsection{Multi-Object Tracking and Referring Multi-Object Tracking}
Multi-object tracking (MOT) aims to maintain consistent object identities across video frames. Early methods follow the tracking-by-detection paradigm \cite{bewley2016simple,wojke2017simple,zhang2022bytetrack,cao2023observation,du2023strongsort}, linking detections over time using motion and appearance cues. Joint detection-and-tracking frameworks \cite{zhang2021fairmot,chen2024delving} further integrate localization and association end-to-end for better spatial-temporal consistency, while Transformer-based trackers \cite{meinhardt2022trackformer,zeng2022motr,yu2023motrv3,li2024matching,li2025ovtr,gao2025multiple} enhance temporal reasoning and global context modeling through self-attention. To further align tracking with natural language, Referring Multi-Object Tracking (RMOT) has been proposed, enabling semantically aware association guided by language descriptions. End-to-end methods such as TransRMOT \cite{wu2023referring}, TempRMOT \cite{zhang2024bootstrapping}, and ROMOT \cite{li2025romot} jointly optimize language grounding and tracking in a unified framework. Two-stage methods including iKUN \cite{du2024ikun}, ReferGPT \cite{chamiti2025refergpt}, and CDRMT \cite{liang2025cognitive} first generate trajectories and then match them with descriptions. CRMOT \cite{chen2025cross}, DRMOT \cite{chen2026drmot}, and RT-RMOT \cite{yu2026rt} further extend RMOT to cross-view, RGBD, and RGB-Thermal settings, demonstrating the growing diversity of this field.

\subsection{Omnidirectional Perception}
Omnidirectional perception aims to understand 360° visual data, yet the widely adopted equirectangular projection (ERP) often introduces geometric distortions that complicate object localization and motion estimation. Large-scale datasets such as Leader360V \cite{zhang2025leader360v} further advance this field by providing real-world 360° video resources. Building upon these advances, omnidirectional multi-object tracking has emerged as a key direction: MMPAT \cite{he2021know} enhances detection and trajectory inference by fusing 2D omnidirectional views with 3D point clouds, while OmniTrack \cite{luo2025omnidirectional} integrates Tracklet Management with CircularStatE for geometric-aware temporal reasoning, achieving state-of-the-art performance on omnidirectional tracking benchmarks.

\subsection{Large Vision-Language Models}
Large Vision-Language Models (LVLMs) have evolved from vision–language alignment to multimodal reasoning. Early efforts such as CLIP \cite{radford2021learning} established image–text alignment, while GPT-4 \cite{achiam2023gpt} extended multimodal understanding through unified perception and reasoning. Recent LVLMs emphasize video comprehension and temporal grounding: InternVL3.5 \cite{wang2025internvl3} advances scalable multimodal learning, and Qwen2.5-VL \cite{bai2025qwen2} enhances fine-grained visual grounding, making it particularly suitable for language-guided detection tasks. Florence2 \cite{xiao2024florence} further pushes the boundaries of efficient vision–language modeling and instruction tuning. ChatTracker \cite{sun2024chattracker} leverages multimodal large language models to enhance tracking performance, demonstrating the growing potential of LVLMs in tracking applications.

\section{Dataset}
\label{sec:Dataset}

\subsection{Dataset Collection}
Our dataset is built upon the JackRabbot Dataset and Benchmark (JRDB) \cite{martin2019jrdb}, which provides omnidirectional capture with complete 360° coverage of the surrounding environment, along with per-frame person IDs and bounding box annotations. However, it lacks high-level semantic descriptions necessary for language-guided tasks. To address this, we introduce ORSet, which augments JRDB with comprehensive language descriptions capturing both spatial and temporal semantics, enabling language-driven multimodal research under omnidirectional perception settings.

\begin{figure*}[t] 
    \centering
    \includegraphics[width=1.0\linewidth]{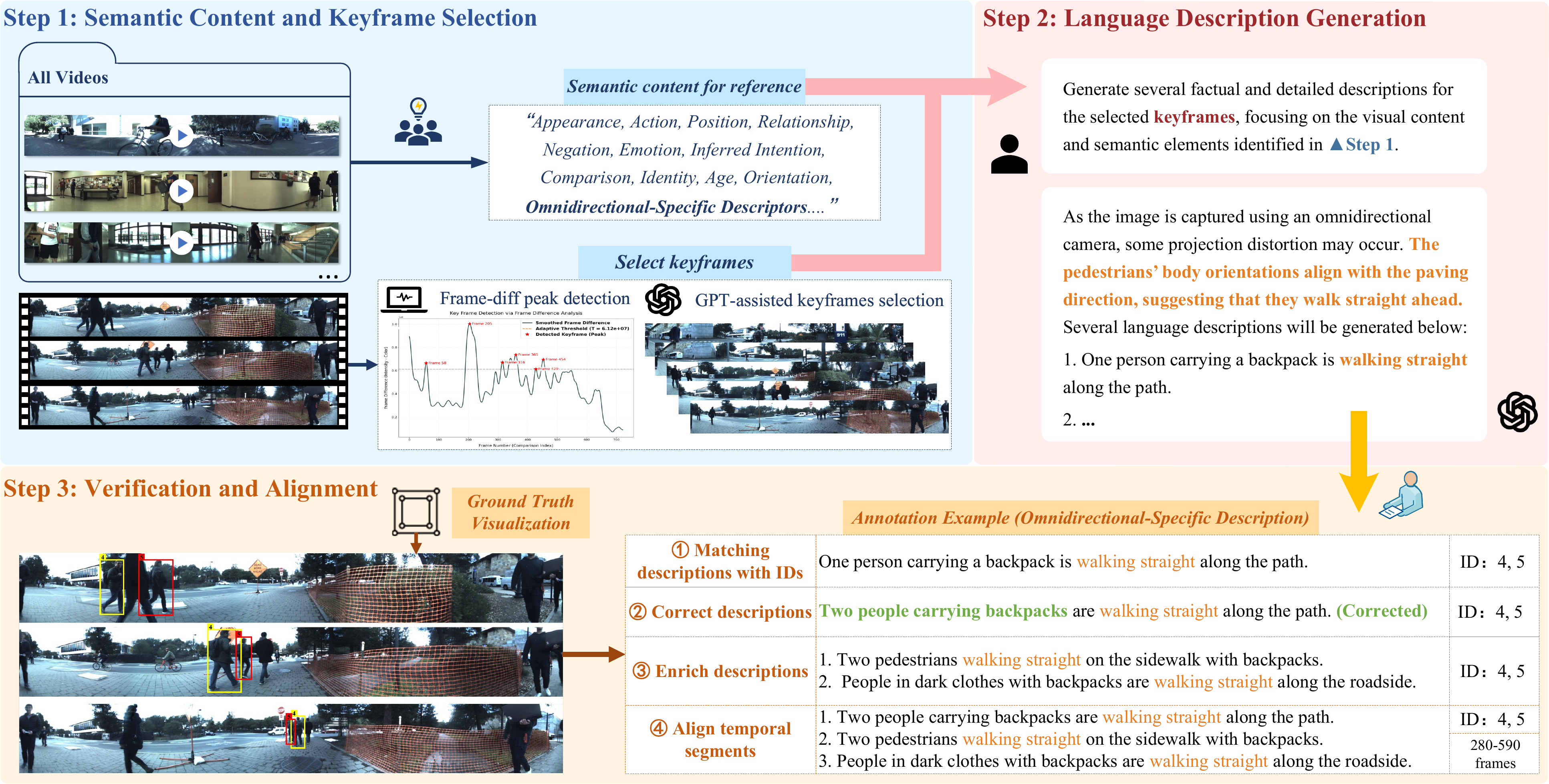}
    \caption{\textbf{Overview of the multi-stage language description annotation pipeline.} \textbf{Step 1:} Semantic content selection and representative keyframe identification via algorithmic detection and GPT-assisted refinement. \textbf{Step 2:} GPT-4o generates diverse descriptions based on selected keyframes and predefined content. \textbf{Step 3:} Human annotators match descriptions to ground-truth person IDs, verify accuracy, enrich linguistic expressions, and align them with temporal segments.}
    \label{fig:Overview of the multi-stage language description annotation pipeline.} 
\end{figure*}

\subsection{Dataset Annotation}
To ensure high-quality and well-aligned language description annotations, we design a systematic three-stage annotation pipeline, as illustrated in \cref{fig:Overview of the multi-stage language description annotation pipeline.}.

\textbf{Step 1: Semantic Content and Keyframe Selection.}
We systematically review all videos to determine what to describe, including each person's appearance, movements or postures, spatial relationships with the environment, social interactions, and the distinctive trajectories captured by omnidirectional cameras. Representative keyframes are identified using frame-difference peak detection, where local maxima exceeding an adaptive threshold are selected as candidates representing moments of significant motion or scene variation, with consecutive nearby peaks merged to avoid redundancy. A GPT-assisted refinement step further filters these candidates by evaluating whether each keyframe captures a semantically meaningful event, such as a direction change, occlusion onset, or interaction, rather than merely a frame with large pixel-level difference.

\textbf{Step 2: Language Description Generation.}
We employ GPT-4o \cite{achiam2023gpt} to generate candidate descriptions based on sampled keyframes and semantic references from Step 1. A fixed prompt template (as shown in \cref{fig:Overview of the multi-stage language description annotation pipeline.}, Step 2) is used to ensure consistency. For each trajectory, GPT-4o generates multiple candidate descriptions focusing on the individual and their motion patterns within the scene.

\textbf{Step 3: Verification and Alignment.}
Human annotators match each generated description with its corresponding ground-truth trajectory ID visualized in the video. All descriptions are then manually refined and temporally aligned with the trajectory spans to ensure semantic accuracy and consistency across frames. Finally, a thorough verification pass checks correctness and enriches linguistic expressions for clarity and coherence.

\begin{figure}[!t]
    \centering
    \includegraphics[width=1.0\linewidth]{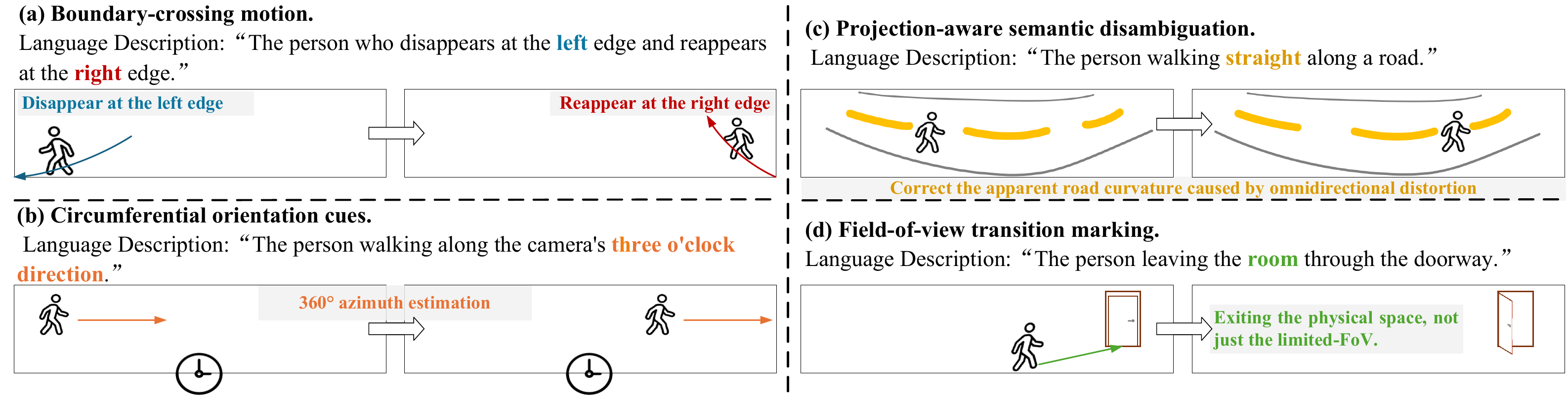}
    \caption{\textbf{Visualization of the four omnidirectional-specific descriptors.} (a) \textbf{Boundary-crossing motion:} links objects disappearing and reappearing across panoramic seams. (b) \textbf{Circumferential orientation cues:} uses a 360° coordinate system for globally consistent direction. (c) \textbf{Projection-aware semantic disambiguation:} infers true motion direction from geometric cues despite projection distortion. (d) \textbf{Field-of-view transition marking:} distinguishes physical exits from view limitations.}
    \label{fig:Visualization of the four omnidirectional-specific descriptors.}
\end{figure}

\textbf{Omnidirectional-Specific Descriptors.}
Omnidirectional imagery exhibits two prominent characteristics: (i) full 360$^\circ$ field-of-view coverage, which enables motion and exit behaviors to be described through coordinate- and orientation-aware cues; and (ii) projection-induced distortion and stitching artifacts, which may introduce misleading visual patterns. Based on these characteristics, we categorize omnidirectional-specific descriptions into four types, as illustrated in \cref{fig:Visualization of the four omnidirectional-specific descriptors.}. These descriptors may be annotated individually or in combination, reflecting overlapping semantics in real scenes. They complement conventional referring expressions and enable ORSet to evaluate both general referring ability and omnidirectional-specific challenges.

\noindent (a) \textbf{\textit{Boundary-crossing motion.}} When an omnidirectional view is unfolded, seams may cause continuous motion to appear as if an object disappears on one side and reappears on the other. We explicitly describe this phenomenon to help the model recognize it as a single continuous 3D motion (e.g., “A person disappears at the left edge and reappears at the right edge.“) rather than two separate targets.

\noindent (b) \textbf{\textit{Circumferential orientation cues.} }By leveraging the 360° coordinate system, we express directions precisely using clock-position notation (e.g., “at the 12 o'clock position"). This camera-centered anchoring provides globally consistent orientation information, overcoming the ambiguity of relative expressions such as “left" or “right".

\noindent (c) \textbf{\textit{Projection-aware semantic disambiguation.}} Omnidirectional projection may cause structurally continuous elements (e.g., roads or building edges) to appear curved or stretched. During annotation, we leverage environmental geometric cues such as lane markings and shadow directions to guide annotators in inferring the true motion direction of targets. The original images remain unchanged throughout.

\noindent (d) \textbf{\textit{Field-of-view transition marking.}} In omnidirectional scenes, objects may enter or exit the 360° field of view from any direction. We explicitly annotate such events (e.g., “A person leaving the room through the doorway.") to help the model build a coherent understanding of scene dynamics.

\subsection{Dataset Split}
We split all video sequences into training and test sets at a 6:4 ratio, resulting in 17 training scenes and 10 test scenes. Language descriptions, person identities, and location information are assigned to the corresponding subsets accordingly.

\begin{table}[t]
    \centering
    \captionsetup{font=small, labelfont=bf} 
    
    \begin{minipage}[t]{0.43\textwidth}
        \centering
        \vspace{0pt} 
        \captionof{table}{Statistics of the ORSet dataset.}
        \label{tab:Statistics of the ORSet dataset.}
        \resizebox{\linewidth}{!}{
            \begin{tabular}{lc}
                \toprule
                \textbf{Statistic} & \textbf{Value} \\
                \midrule
                Video Scenes & 27 \\
                Total BBoxes & 2.9 M \\
                Language Descriptions & 848 \\
                Omnidirectional Descriptions & 175 \\
                Avg. Frames / Desc. & 575.3 \\
                Avg. Desc. Length & 8.2 words \\
                Camera State & 13 / 14 \\
                Image Size & 3760 $\times$ 480 \\
                \bottomrule
            \end{tabular}
        }
    \end{minipage}
    \hfill 
    \begin{minipage}[t]{0.55\textwidth}
        \centering
        \vspace{0pt} 
        \captionof{lstlisting}{Annotation format of the ORSet dataset.}
        \label{fig:ORSet_format.}
        \begin{lstlisting}[
            style=mystyle,
            breaklines=true,
            % 代码字号也稍微调小，防止溢出
            basicstyle=\fontsize{6.5}{7.5}\selectfont\ttfamily,
            aboveskip=3pt,
            belowskip=0pt,
            frame=single % 加上边框，看起来更整齐
        ]
{
  "scene_id": "bytes-cafe-2019...",
  "frame_range": [[120,450], [600,860]],
  "annotations": {
    "120": [{"id": 3, "bbox": [x1,y1,x2,y2]}],
    "121": [
      {"id": 3, "bbox": [x1,y1,x2,y2]},
      {"id": 8, "bbox": [x1,y1,x2,y2]}
    ]
  },
  ...
  "description": "The person who orders food."
}
        \end{lstlisting}
    \end{minipage}
\end{table}

\begin{figure*}[t]
\begin{subfigure}{0.25\linewidth} 
  \raggedright 
  \includegraphics[width=0.95\linewidth]{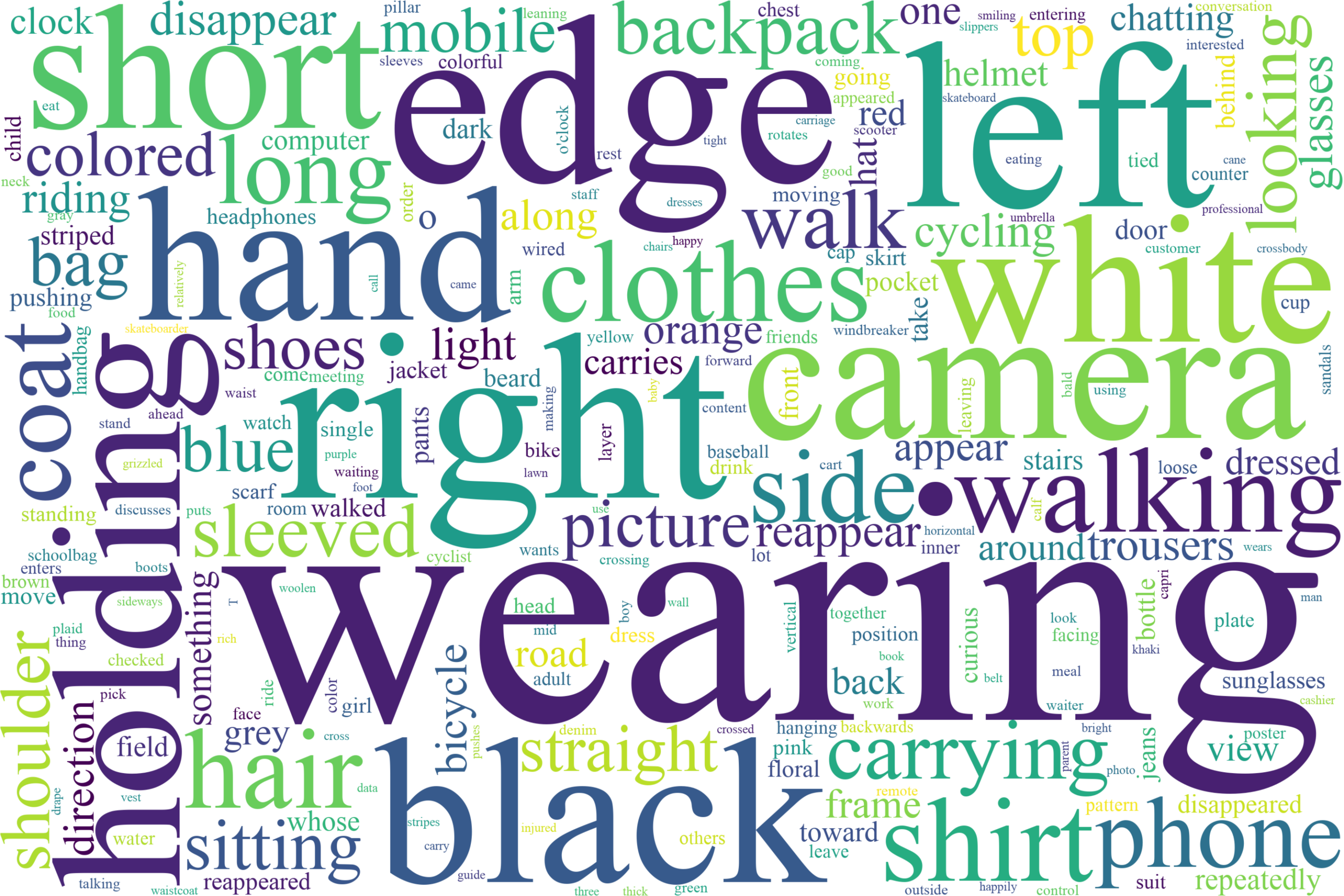}
  \caption{}
  \label{fig:a.}
\end{subfigure}
\hfil
\begin{subfigure}{0.24\linewidth} 
  \centering
  \includegraphics[width=0.86\linewidth]{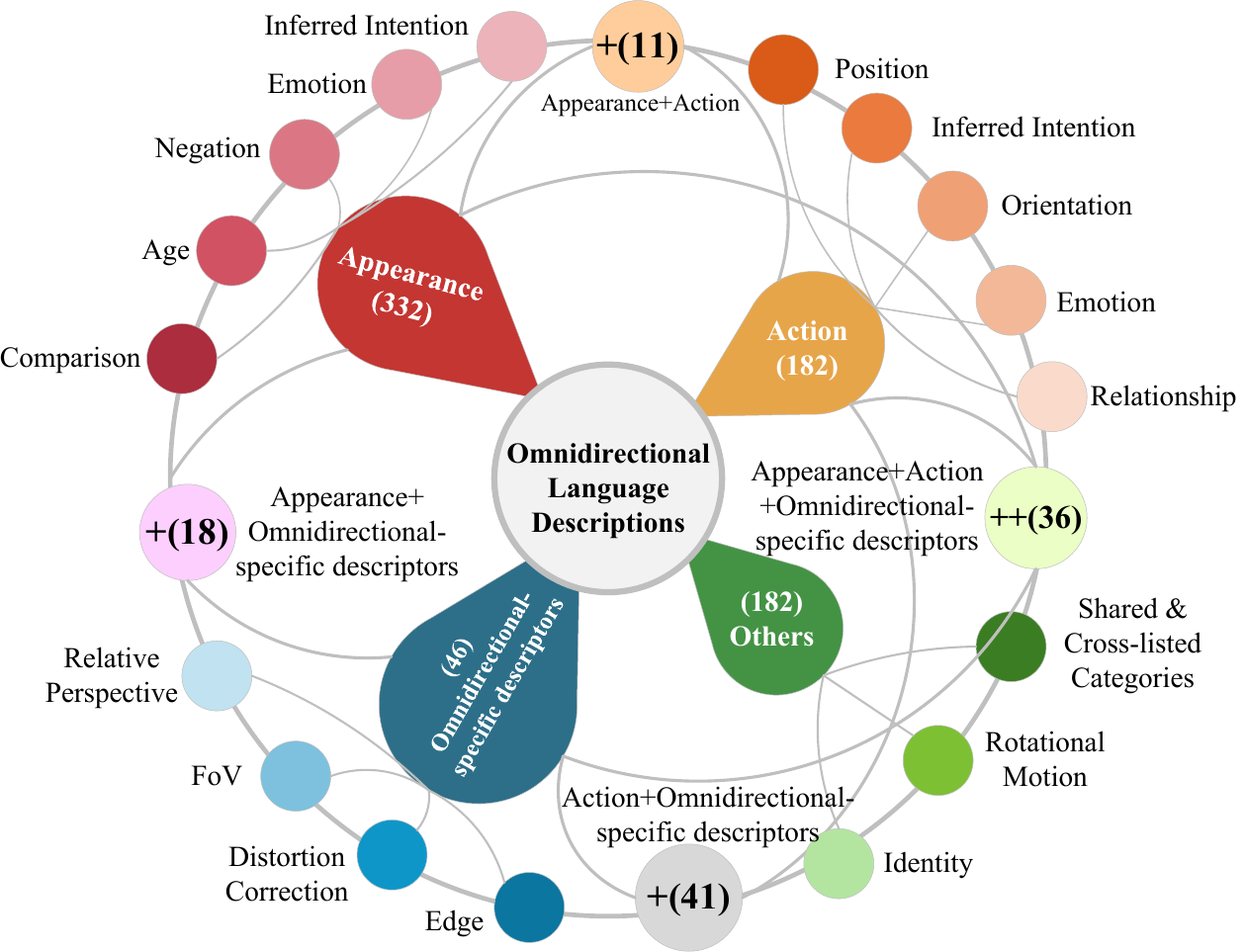}
  \caption{}
  \label{fig:b}
\end{subfigure}
\hfil
\begin{subfigure}{0.25\linewidth} 
  \centering
  \includegraphics[width=1.0\linewidth]{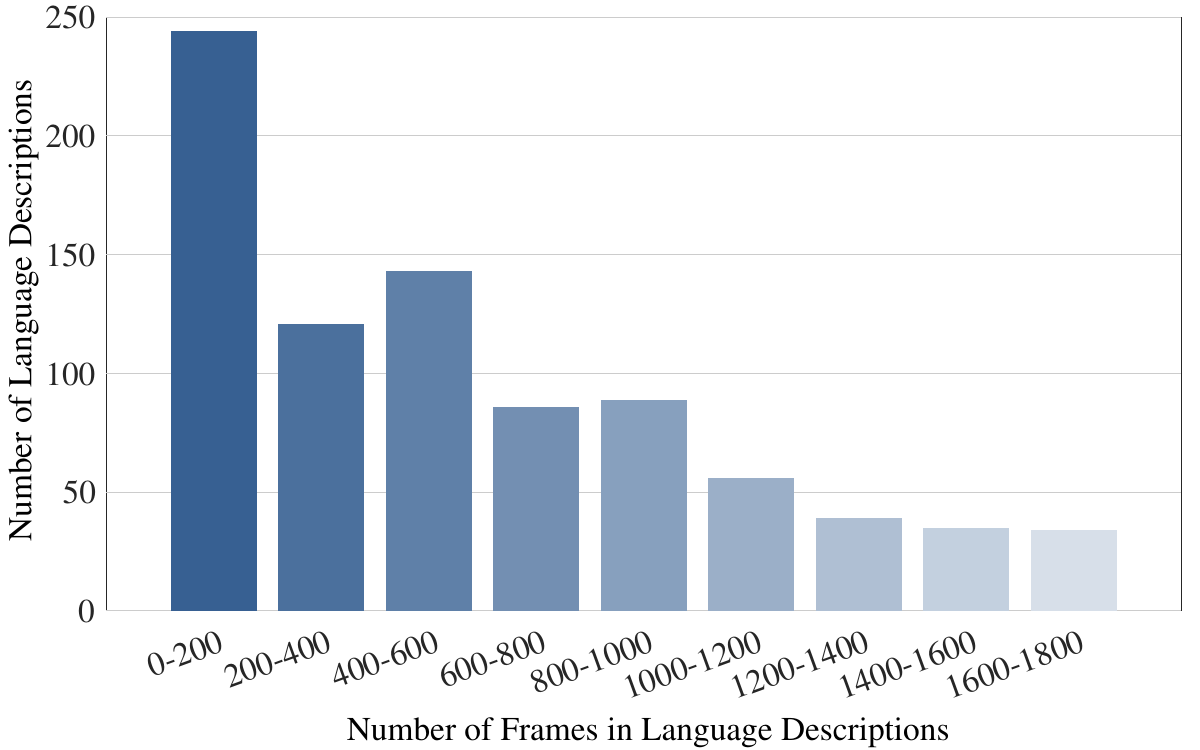}
  \caption{}
  \label{fig:c}
\end{subfigure}
\hfil
\begin{subfigure}{0.24\linewidth} 
  \centering
  \includegraphics[width=1.0\linewidth]{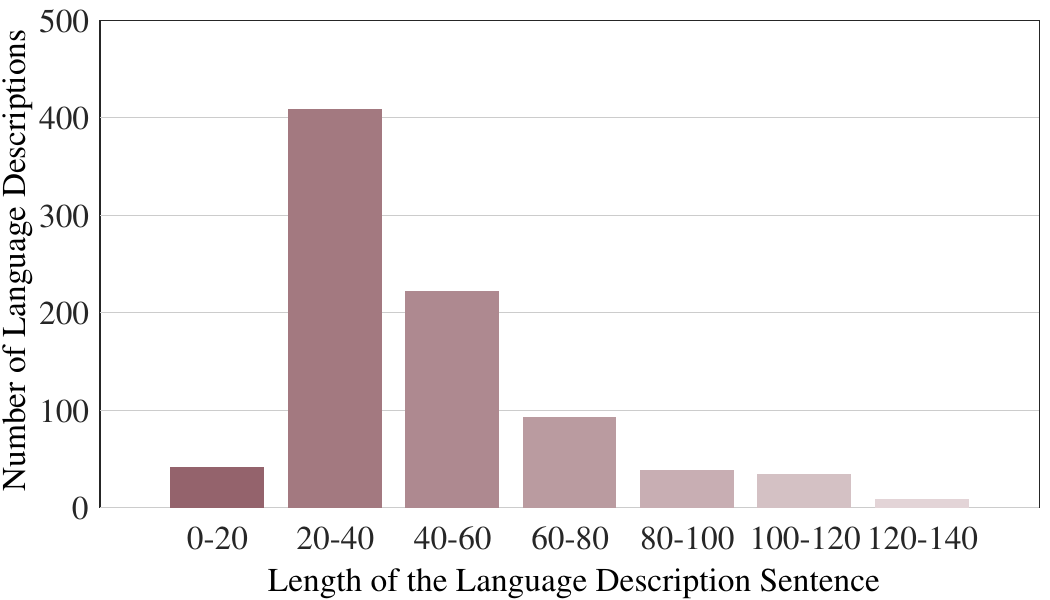}
  \caption{}
  \label{fig:d}
\end{subfigure}
\caption{\textbf{Overview of the ORSet dataset statistics.} (a) \textbf{Word Cloud:} frequent terms cover appearance (e.g., “black"), actions (e.g., “walking"), and omnidirectional-specific cues (e.g., “edge"). (b) \textbf{Description Types:} most descriptions blend multiple attributes. (c) \textbf{Description Lengths:} majority fall in the 20–80 character range. (d) \textbf{Track Lengths:} balanced short- and long-term coverage supports temporal reasoning.}
\label{fig:Overview of the ORSet dataset statistics.}
\end{figure*}

\subsection{Dataset Statistics}
The detailed statistics of ORSet are summarized in \cref{tab:Statistics of the ORSet dataset.}, and the annotation format is illustrated in \cref{fig:ORSet_format.}.

\textbf{Word Cloud.}
As illustrated in \cref{fig:Overview of the ORSet dataset statistics.} (a), prominent terms cover appearance (e.g., “black", “shirt"), actions (e.g., “walking", “carrying"), and omnidirectional-specific cues (e.g., “edge"). The high frequency of “edge" reflects the inclusion of Omnidirectional-Specific Descriptors, highlighting the dataset's distinction from general vision–language datasets.

\textbf{Distribution of Language Description Types.}
As depicted in \cref{fig:Overview of the ORSet dataset statistics.} (b), descriptions are categorized into three main groups: Appearance (332), Action (182), and Omnidirectional-specific descriptors (46), where counts represent descriptions belonging solely to each type. Many descriptions combine multiple categories, as indicated by the “+" nodes, reflecting real-world language patterns that integrate multiple attributes rather than relying on a single feature.

\textbf{Distribution of Language Description Lengths.}
As shown in \cref{fig:Overview of the ORSet dataset statistics.} (c), most descriptions fall between 20 and 80 characters, with a smaller portion exceeding 100 characters. This distribution reflects natural variability in human expression while remaining consistent with everyday spoken language, making the dataset well-suited for studying natural language–vision alignment.

\textbf{Distribution of Track Lengths.}
As shown in \cref{fig:Overview of the ORSet dataset statistics.} (d), most descriptions correspond to short sequences within 0–200 frames, with a gradual decrease as frame range increases. Medium-length descriptions (200–1000 frames) maintain a moderate presence, while sequences exceeding 1000 frames are relatively rare. This long-tailed distribution provides balanced coverage of both short-term and long-term temporal contexts, supporting fine-grained temporal grounding in the ORMOT setting.

\section{Methodology}
\label{sec:Methodology}

\subsection{Overview}

\begin{figure*}[t]
    \centering
    \includegraphics[width=1.0\linewidth]{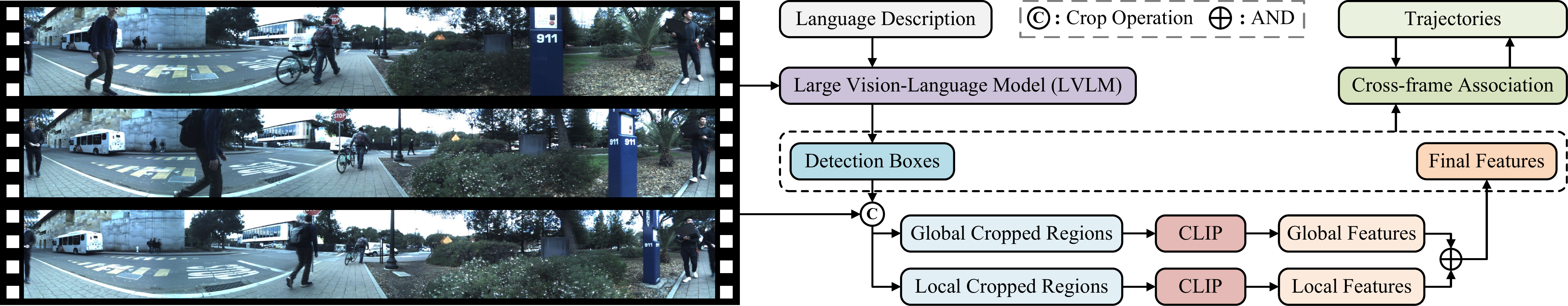}
    \caption{\textbf{Pipeline of ORTrack}, comprising three components: (1) \textbf{Language-guided detection:} an LVLM serves as an open-vocabulary detector to output bounding boxes conditioned on the language description. (2) \textbf{Two-stage cropping-based feature extraction:} hierarchical region extraction and encoding to obtain discriminative features. (3) \textbf{Cross-frame association:} linking detected boxes via cosine similarity and Hungarian matching to maintain consistent identities.}
    \label{fig:Pipeline of ORTrack.}
\end{figure*}

Given a sequence of omnidirectional frames $\left\{I_t\right\}_{t=1}^T$ and a language description $L$, ORTrack produces a set of trajectories $\mathcal{T}=\left\{\tau_k\right\}_{k=1}^K$, where each $\tau_k=\left\{b_t^k\right\}_{t=t_s}^{t_e}$ represents a continuous tracklet corresponding to a language-referred target. The pipeline of our ORTrack framework is shown in \cref{fig:Pipeline of ORTrack.}.

\subsection{Language-guided Detection via LVLM}
Omnidirectional environments cover 360° fields of view, often containing multiple entities with varying appearances and scales. Conventional detectors suffer from limited predefined classes and lack flexibility to respond to language descriptions. To address this, ORTrack employs an LVLM-based detection mechanism that integrates visual grounding and semantic understanding in a unified reasoning process. Given a frame $I_t$ and a text instruction $L$, the LVLM (e.g., Qwen2.5-VL \cite{bai2025qwen2}, InternVL3.5 \cite{wang2025internvl3}) aligns visual and language representations to predict bounding boxes for referred targets. Formally:
\begin{equation}
\left\{b_t^i\right\}_{i=1}^{N_t}=\operatorname{LVLM}\left(I_t, L\right)=\operatorname{Align}\left(\psi_v\left(I_t\right), \psi_l(L)\right)
\end{equation}
where $b_t^i=\left(x_t^i, y_t^i, w_t^i, h_t^i\right)$ represents the bounding box of the $i$-th referred object.

\subsection{Two-stage Cropping-based Feature Extraction}
Following detection, ORTrack performs two-stage cropping-based inference to obtain robust object features, balancing global contextual cues and local discriminative details. Given that omnidirectional images are in equirectangular projection, where objects may span from the leftmost to the rightmost boundary, IoU-based matching becomes unreliable. ORTrack therefore employs a feature-based association strategy.

\textbf{Stage 1: Global Contextual Cropping.}
Omnidirectional frames often suffer from wide-angle compression, which can destabilize local features. To mitigate this effect, each detected box is expanded by a margin ratio $\alpha$ to incorporate surrounding contextual information:
\begin{equation}
I_t^{i, \text { global }}=\operatorname{Crop}\left(I_t, \alpha b_t^i\right)
\label{eq:I_global.}
\end{equation}

where $I_t^{i, \text { global }}$ represents the global cropped region of the $i$-th referred object. The “Crop($\cdot$)" represents the cropping operation. $I_t$ represents the frame. $\alpha$ represents the margin ratio.

\textbf{Stage 2: Fine-grained Target Cropping.}
The second-stage crop extracts the exact target region $I_t^{i, \text { local }}$, defined by the detected bounding box $b_t^i$:
\begin{equation}
I_t^{i, \text { local }}=\operatorname{Crop}\left(I_t, b_t^i\right)
\end{equation}

Both global and local regions are processed by a frozen CLIP \cite{radford2021learning} visual encoder to generate features:
\begin{equation}
\mathbf{f}_{t, \text { global }}^i=\phi_g\left(I_t^{i, \text { global }}\right)
\end{equation}
\begin{equation}
\mathbf{f}_{t, \text { local }}^i=\phi_l\left(I_t^{i, \text { local }}\right)
\end{equation}
where $\mathbf{f}_{t, \text { global }}^i$ represents the global feature of the $i$-th referred object. $\mathbf{f}_{t, \text { local }}^i$ represents the local feature of the $i$-th referred object. $\phi$ denotes the CLIP visual encoder.

The final object feature is computed as:
\begin{equation}
\mathbf{f}_t^i = \mathbf{f}_{t, \text { local }}^i + \lambda \mathbf{f}_{t, \text { global }}^i
\label{eq:final object feature.}
\end{equation}
where $\lambda$ is the feature fusion weight.

\subsection{Cross-frame Association}

To maintain identity consistency, ORTrack adopts cosine-similarity-based matching followed by Hungarian algorithm \cite{kuhn1955hungarian}. Given two consecutive frames $t$ and $t+1$, we compute the pairwise cosine similarity to construct a similarity matrix:
\begin{equation}
S_{i j}=\frac{\mathbf{f}_t^i \cdot \mathbf{f}_{t+1}^j}{\left\|\mathbf{f}_t^i\right\|\left\|\mathbf{f}_{t+1}^j\right\|}
\end{equation}

The corresponding cost matrix $C_{ij} = 1 - S_{ij}$ is used to find the optimal assignment:
\begin{equation}
\min _X \sum_{i, j} C_{i j} X_{i j}, \quad \text { s.t. } X_{i j} \in\{0,1\}.
\end{equation}
where $X_{i j}=1$ indicates that detection $j$ in frame $t+1$ is assigned to track $i$ in frame $t$. Unmatched detections initiate new tracklets, while tracks without matches for $\tau_{\max }$ frames are terminated.

\section{Experiments}
\label{sec:Experiments}

\subsection{Metrics}
We follow the standard evaluation metrics \cite{wu2023referring}, including HOTA, DetA, AssA, DetRe, DetPr, AssRe, AssPr, and LocA, which jointly assess detection accuracy, association consistency, and localization precision.

\subsection{Implementation Details}
ORTrack adopts Qwen2.5-VL-7B \cite{bai2025qwen2} as the LVLM backbone and CLIP-ViT-B-32 \cite{radford2021learning} as the feature encoder. The margin ratio $\alpha$ in \cref{eq:I_global.} is set to 1.2 and the fusion weight $\lambda$ in \cref{eq:final object feature.} is set to 0.5. All experiments are conducted on a single NVIDIA RTX A6000 GPU. 

To ensure fair cross-model comparison, we unify the prompt construction strategy while allowing minor adaptations for each model's instruction format. For DeepSeek-VL-7B \cite{lu2024deepseek}, LLaVA-NEXT-8B \cite{liu2024llavanext}, and InternVL3.5-8B \cite{wang2025internvl3}, the prompt asks the model to “Please detect all \textit{\{description\}} in the image and output their coordinates in the [x1, y1, x2, y2] format." For the Qwen2.5-VL series \cite{bai2025qwen2}, the prompt asks the model to “Please detect and label all \textit{\{description\}} in the following image and mark their positions."

\subsection{Quantitative Results}
Since ORSet is a newly constructed dataset, we restrict comparison to open-source RMOT methods that can be directly evaluated without additional access constraints. As shown in \cref{tab:Performance of different methods on the test set of the ORSet dataset under zero-shot conditions.}, ORTrack consistently outperforms prior works across most metrics. Notably, ORTrack achieves substantial improvements in HOTA (9.97 vs. 2.41/2.00), DetA (6.37 vs. 1.40/0.45), and AssA (16.15 vs. 4.24/9.01), with balanced gains in both precision and recall metrics, demonstrating strong detection and association performance under zero-shot conditions.

\begin{table*}[t]
\centering
\caption{\textbf{Performance of different methods} on the test set of the ORSet dataset under zero-shot conditions. ↑ indicates that higher score is better. The best results are marked in \textbf{bold}.}
\label{tab:Performance of different methods on the test set of the ORSet dataset under zero-shot conditions.}
\resizebox{1.0 \linewidth}{!}{
    \begin{tabular}{lccccccccc}
        \toprule
        Method & Published & HOTA↑ & DetA↑ & AssA↑ & DetRe↑ & DetPr↑ & AssRe↑ & AssPr↑ & LocA↑ \\
        \midrule
        TransRMOT \cite{wu2023referring} & CVPR 2023 & 2.41 & 1.40 & 4.24 & 1.56 & 12.10 & 4.29 & 79.29 & 75.05 \\
        TempRMOT \cite{zhang2024bootstrapping} & ArXiv  2024 & 2.00 & 0.45 & 9.01 & 0.47 & 11.01 & 9.44 & \textbf{80.93} & 78.88 \\
        
        \midrule
        \textbf{ORTrack (Ours)} & - & \textbf{9.97} & \textbf{6.37} & \textbf{16.15} & \textbf{9.20} & \textbf{16.69} & \textbf{17.35} & 61.80 & \textbf{79.68} \\
        \bottomrule
    \end{tabular}
}
\end{table*}

\begin{table*}[t]
\centering
\caption{\textbf{Ablation study} of our ORTrack framework on the test set of the ORSet dataset under zero-shot conditions. The \textcolor{orange}{number} in the first column indicates the row number. ↑ indicates that higher score is better. The best results are marked in \textbf{bold}.}
\label{tab:Ablation study of our ORTrack framework on the test set of the ORSet dataset under zero-shot conditions.}
\resizebox{1.0 \linewidth}{!}{
    \setlength{\tabcolsep}{2.2mm}{
        \begin{tabular}{clccccccccc}
            \toprule
            & \multicolumn{2}{c}{Large Vision-Language Models (LVLMs)} & \multicolumn{8}{c}{Metrics} \\
            \cmidrule(lr){2-3} \cmidrule(lr){4-11}
            \rule{0pt}{10pt} & Name & Size & HOTA↑ & DetA↑ & AssA↑ & DetRe↑ & DetPr↑ & AssRe↑ & AssPr↑ & LocA↑ \\
            \midrule
            \textcolor{orange}{1} &  DeepSeek-VL \cite{lu2024deepseek} & 7B & 0.12 & 0.06 & 0.29 & 0.06 & 2.83 & 0.30 & 17.89 & 66.66 \\
            \textcolor{orange}{2} & LLaVA-NEXT \cite{liu2024llavanext} & 8B & 1.76 & 1.11 & 3.12 & 1.55 & 3.35 & 3.80 & 16.22 & 57.57 \\
            \textcolor{orange}{3} & InternVL3.5 \cite{wang2025internvl3} & 8B & 0.55 & 0.49 & 0.72 & 0.52 & 6.93 & 0.74 & 39.48 & 61.56 \\
            \midrule
            \textcolor{orange}{4} & Qwen2.5-VL \cite{bai2025qwen2} & 7B & \textbf{9.97} & \textbf{6.37} & \textbf{16.15} & \textbf{9.20} & \textbf{16.69} & \textbf{17.35} & \textbf{61.80} & \textbf{79.68} \\
            \textcolor{orange}{5} & Qwen2.5-VL \cite{bai2025qwen2} & 3B & 3.33 & 2.13 & 5.84 & 2.72 & 8.54 & 6.08 & 52.44 & 71.34 \\
            \bottomrule
        \end{tabular}
    }
}
\end{table*}

\begin{table*}[t]
 \begin{center}
  \caption{\textbf{Efficiency and performance of different feature encoder.} ↑ indicates that higher score is better. ↓ indicates that lower score is better. The best results are marked in \textbf{bold}.}
  \label{tab:Efficiency and performance of different feature encoder.}
  \setlength{\tabcolsep}{2.15pt} 
  \resizebox{1.0 \linewidth}{!}
  {
    \begin{tabular}{l cc cc cccccccc}
        \toprule
        \multirow{2}{*}{Method} & \multicolumn{2}{c}{Feature} & \multicolumn{2}{c}{Efficiency} & \multicolumn{8}{c}{Metrics} \\
        \cmidrule(lr){2-3} \cmidrule(lr){4-5} \cmidrule(lr){6-13} 
         & \textbf{Encoder} &  \textbf{Dims} & \textbf{FPS↑} & FLOPs↓ & HOTA↑ & DetA↑ & AssA↑ & DetRe↑ & DetPr↑ & AssRe↑ & AssPr↑ & LocA↑ \\
        \midrule
        w/o CLIP & \textbf{LVLM} & \textbf{3584} & 0.399 & \textbf{4.193$\times10^{13}$} & \textbf{10.05} & \textbf{6.37} & \textbf{16.42} & \textbf{9.20} & 16.68 & \textbf{17.65} & 61.74 & \textbf{79.68} \\
        ORTrack (Ours) & \textbf{CLIP} & \textbf{512} & \textbf{0.446} & 4.194$\times10^{13}$ & 9.97 & \textbf{6.37} & 16.15 & \textbf{9.20} & \textbf{16.69} & 17.35 & \textbf{61.80} & \textbf{79.68} \\
        \bottomrule
    \end{tabular}
    }
 \end{center}
\end{table*}

\begin{table*}[t]
 \begin{center}
  \caption{\textbf{Performance of different association strategies on the ORSet dataset.} ↑ indicates that higher score is better. The best results are marked in \textbf{bold}.}
  \label{tab:Performance of different association strategies on the ORSet dataset.}
  \setlength{\tabcolsep}{3.15pt} 
  \renewcommand{\arraystretch}{0.9} 
    \begin{tabular}{lcccccccccc}
        \toprule
        Method & HOTA↑ & DetA↑ & AssA↑ & DetRe↑ & DetPr↑ & AssRe↑ & AssPr↑ & LocA↑ \\
        \midrule
        LVLM + OC-SORT \cite{cao2023observation} & 5.05 & 2.56 & 10.19 & 2.83 & \textbf{20.76} & 10.66 & \textbf{73.04} & \textbf{80.70} \\
        ORTrack (Ours) & \textbf{9.97} & \textbf{6.37} & \textbf{16.15} & \textbf{9.20} & 16.69 & \textbf{17.35} & 61.80 & 79.68 \\
        \bottomrule
    \end{tabular}
 \end{center}
\end{table*}

\subsection{Ablation Study}
We conduct ablation experiments on the ORSet test set to assess the effectiveness of each component in ORTrack.

\subsubsection{Analysis of Large Vision-Language Models}

As shown in \cref{tab:Ablation study of our ORTrack framework on the test set of the ORSet dataset under zero-shot conditions.}, ORTrack with Qwen2.5-VL-7B (row \textcolor{orange}{4}) achieves the best overall performance, surpassing DeepSeek-VL (row \textcolor{orange}{1}), LLaVA-NEXT (row \textcolor{orange}{2}), and InternVL3.5 (row \textcolor{orange}{3}). Compared to the smaller Qwen2.5-VL-3B (row \textcolor{orange}{5}), the 7B model shows substantial gains, confirming that a larger model with stronger visual-language alignment leads to more accurate language-guided tracking.

\subsubsection{Analysis of CLIP vs LVLM for Feature Encoder} 

As shown in \cref{tab:Efficiency and performance of different feature encoder.}, LVLM-based feature encoding achieves slightly higher accuracy due to higher-dimensional representations, but incurs significant computational overhead during similarity computation. CLIP provides a more efficient alternative: with lower-dimensional features, it achieves faster processing and higher FPS while maintaining competitive accuracy, offering better overall efficiency.

\subsubsection{Analysis of Different Association Strategies}

As shown in \cref{tab:Performance of different association strategies on the ORSet dataset.}, ORTrack significantly outperforms LVLM + OC-SORT \cite{cao2023observation} across all metrics, with both methods using the same LVLM detector. ORTrack achieves higher HOTA (9.97 vs. 5.05), AssA (16.15 vs. 10.19), DetA (6.37 vs. 2.56), and DetRe (9.20 vs. 2.83), confirming the superiority of ORTrack's association strategy for cross-frame object matching.

\begin{figure*}[t]
    \centering
    \begin{subfigure}{0.495\linewidth}
        \centering
        \includegraphics[width=\linewidth]{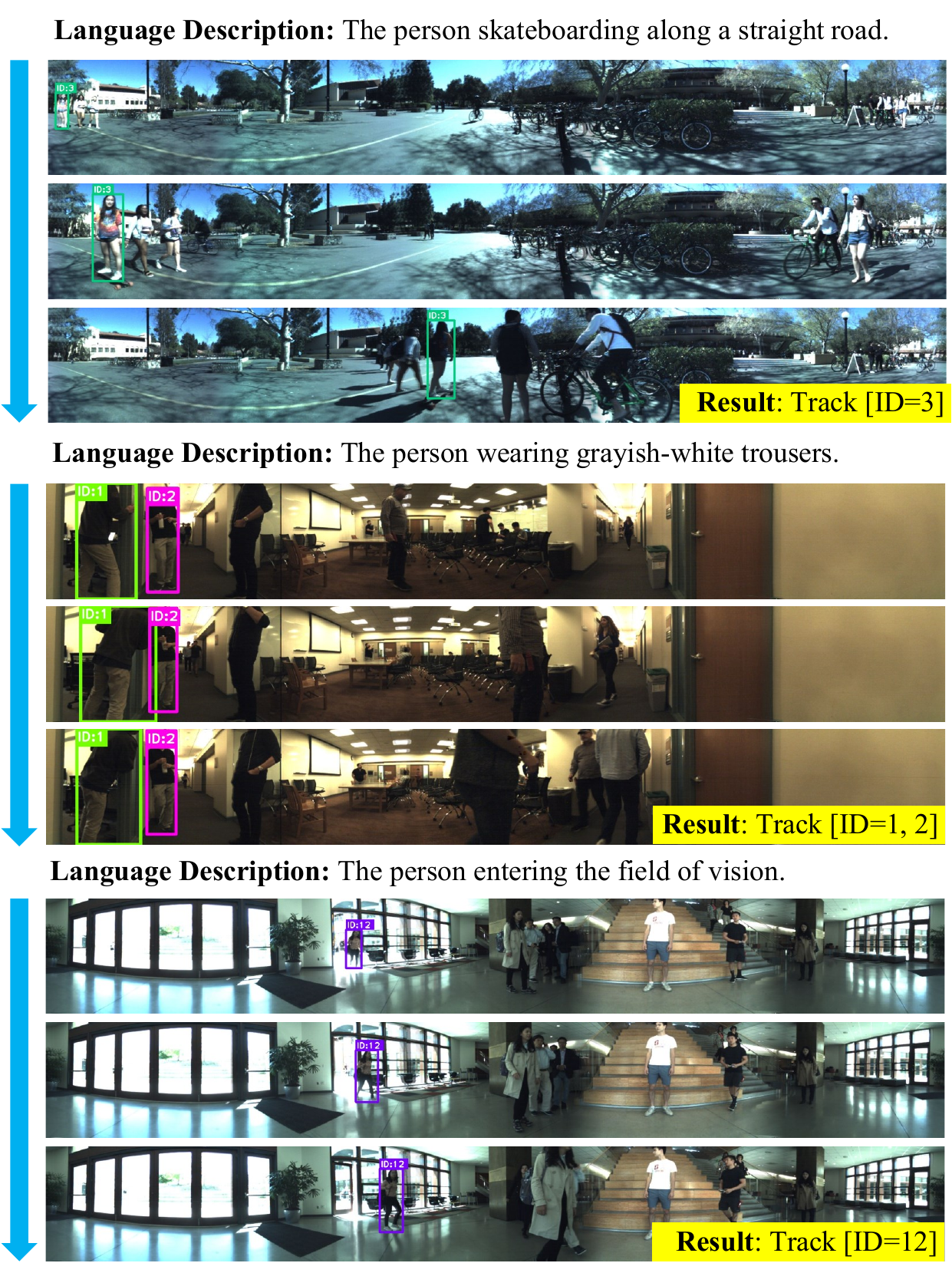}
        \caption{Part 1}
        \label{fig:res1}
    \end{subfigure}
    \hfill
    \begin{subfigure}{0.495\linewidth}
        \centering
        \includegraphics[width=\linewidth]{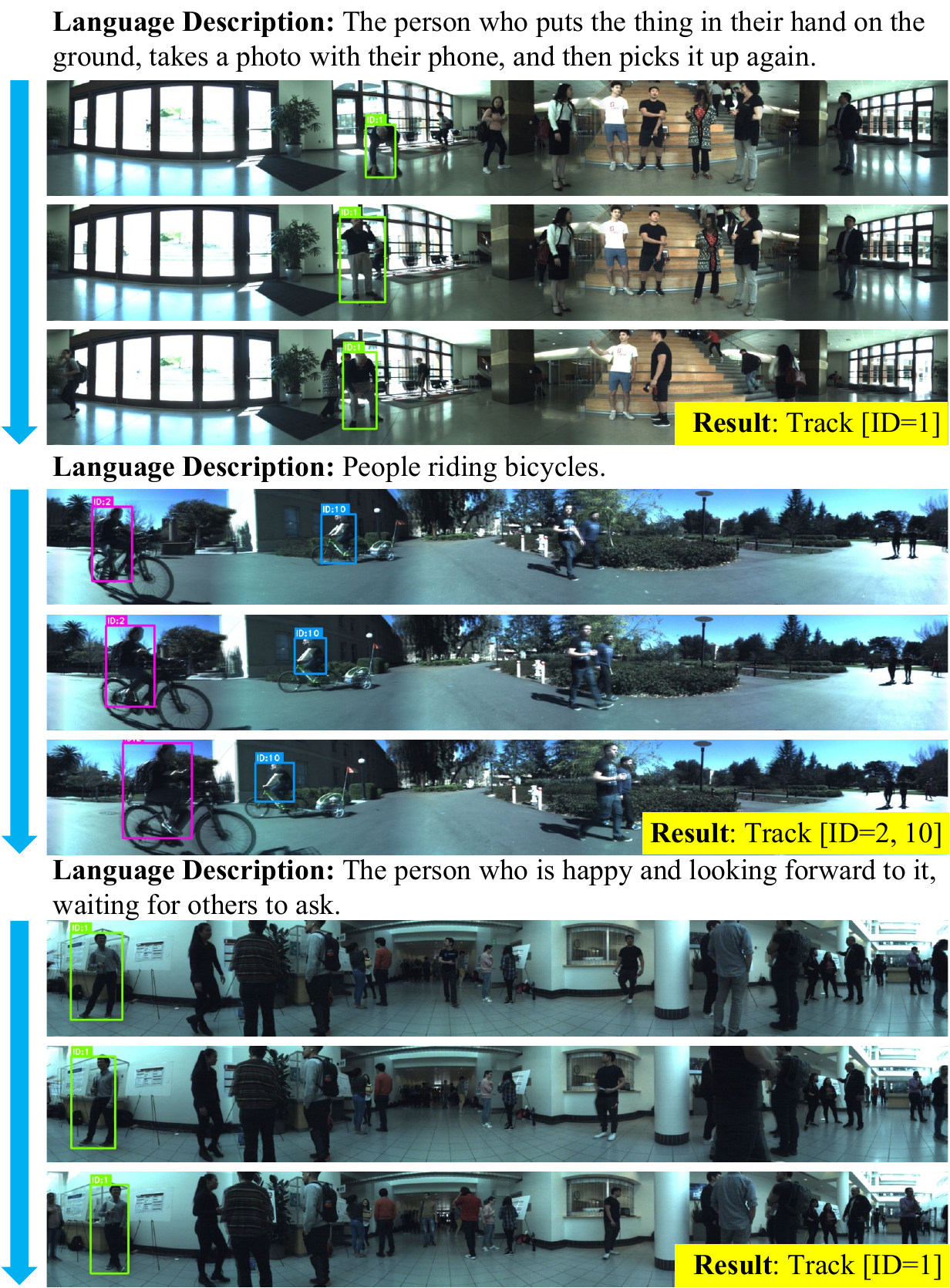}
        \caption{Part 2}
        \label{fig:res2}
    \end{subfigure}
    \caption{\textbf{Qualitative results} of the ORTrack framework under zero-shot conditions. The figure displays representative omnidirectional referring multi-object tracking examples from the test set of the ORSet dataset, divided into Part 1 (a) and Part 2 (b).}
\end{figure*}

\subsection{Qualitative Results}
Qualitative results on the ORSet test set under zero-shot conditions are shown in \cref{fig:res1} and \cref{fig:res2}. The six examples collectively demonstrate ORTrack's capabilities across several challenging scenarios.
Examples (a)1 and (b)2 demonstrate ORTrack's ability to simultaneously track multiple targets (e.g., ID=1,2 and ID=2,10) across frames, validating its multi-object association under omnidirectional settings. In (a)1, despite the road appearing curved due to omnidirectional projection distortion, ORTrack correctly identifies the target described as “skateboarding along a straight road," showing robustness to projection-induced visual distortion. Example (a)3 shows successful tracking of a person described as “entering the field of vision," corresponding to the field-of-view transition descriptor unique to omnidirectional imagery. Examples (b)1 and (b)3 involve complex long-horizon descriptions requiring temporal reasoning and semantic inference, including sequential action understanding (“puts the thing on the ground, takes a photo, and picks it up again") and emotion-based identification (“happy and looking forward to it"), both of which are handled correctly, demonstrating ORTrack's strong language comprehension capability.

\begin{figure}[!t]
    \centering
    \includegraphics[width=1.0\linewidth]{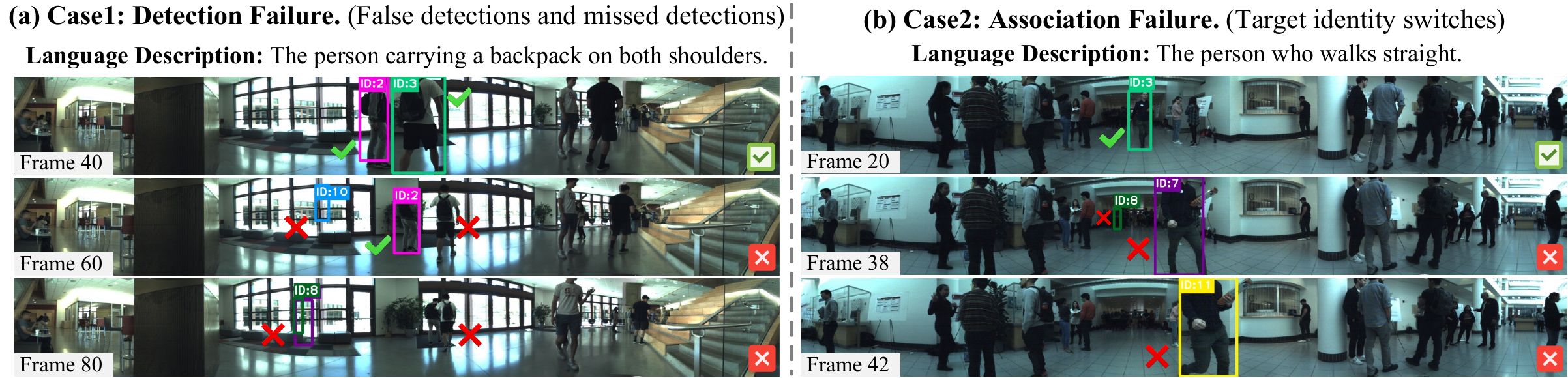}
    \caption{\textbf{Representative Failure Cases of ORMOT.}}
    \label{fig:Failure_case.}
\end{figure}

\subsection{Failure Case Analysis}
Representative failure cases are shown in \cref{fig:Failure_case.}. In Case 1 (a), the system produces false and missed detections, exacerbated by omnidirectional distortion that alters target appearance across the field of view. In Case 2 (b), identity switches occur when targets move in close proximity, causing rapid scale changes and feature drift that disrupt consistent association. These challenges are reflected in \cref{tab:Performance of different association strategies on the ORSet dataset.}, where DetA is generally lower than AssA across all methods, highlighting the difficulty of simultaneous accurate detection and stable association in omnidirectional environments.

\section{Conclusion and Future Work}
\label{sec:Conclusion and Future Work}

In this work, we propose \textbf{Omnidirectional Referring Multi-Object Tracking (ORMOT)}, a novel task that extends RMOT to omnidirectional imagery for understanding long-horizon language descriptions. To facilitate future research, we construct \textbf{ORSet}, comprising 27 omnidirectional scenes, 848 language descriptions, and 3,401 annotated objects, and introduce \textbf{ORTrack}, an LVLM-driven zero-shot baseline that achieves state-of-the-art performance on ORSet.

Future work will focus on improving detection and association robustness under omnidirectional projection distortion, developing more efficient tracking frameworks for practical deployment, and exploring geometry-aware representations tailored to omnidirectional imagery.
\section*{Acknowledgements}

This work was supported in part by Hubei Provincial Science and Technology Plan under Grant 2025BEB006 and in part by the National Natural Science Foundation of China under Grant 62576144.
%
\bibliographystyle{splncs04}
\bibliography{references}

@article{zhang2021fairmot,
  title={Fairmot: On the fairness of detection and re-identification in multiple object tracking},
  author={Zhang, Yifu and Wang, Chunyu and Wang, Xinggang and Zeng, Wenjun and Liu, Wenyu},
  journal={International journal of computer vision},
  volume={129},
  pages={3069--3087},
  year={2021},
  publisher={Springer}
}

@inproceedings{gao2025multiple,
  title={Multiple object tracking as id prediction},
  author={Gao, Ruopeng and Qi, Ji and Wang, Limin},
  booktitle={Proceedings of the Computer Vision and Pattern Recognition Conference},
  pages={27883--27893},
  year={2025}
}

@inproceedings{li2024matching,
  title={Matching Anything by Segmenting Anything},
  author={Li, Siyuan and Ke, Lei and Danelljan, Martin and Piccinelli, Luigi and Segu, Mattia and Van Gool, Luc and Yu, Fisher},
  booktitle={Proceedings of the IEEE/CVF Conference on Computer Vision and Pattern Recognition},
  pages={18963--18973},
  year={2024}
}

@inproceedings{wu2023referring,
  title={Referring multi-object tracking},
  author={Wu, Dongming and Han, Wencheng and Wang, Tiancai and Dong, Xingping and Zhang, Xiangyu and Shen, Jianbing},
  booktitle={Proceedings of the IEEE/CVF conference on computer vision and pattern recognition},
  pages={14633--14642},
  year={2023}
}

@inproceedings{bewley2016simple,
  title={Simple online and realtime tracking},
  author={Bewley, Alex and Ge, Zongyuan and Ott, Lionel and Ramos, Fabio and Upcroft, Ben},
  booktitle={2016 IEEE international conference on image processing (ICIP)},
  pages={3464--3468},
  year={2016},
  organization={IEEE}
}

@inproceedings{wojke2017simple,
  title={Simple online and realtime tracking with a deep association metric},
  author={Wojke, Nicolai and Bewley, Alex and Paulus, Dietrich},
  booktitle={2017 IEEE international conference on image processing (ICIP)},
  pages={3645--3649},
  year={2017},
  organization={IEEE}
}

@inproceedings{zhang2022bytetrack,
  title={Bytetrack: Multi-object tracking by associating every detection box},
  author={Zhang, Yifu and Sun, Peize and Jiang, Yi and Yu, Dongdong and Weng, Fucheng and Yuan, Zehuan and Luo, Ping and Liu, Wenyu and Wang, Xinggang},
  booktitle={European Conference on Computer Vision},
  pages={1--21},
  year={2022},
  organization={Springer}
}

@article{du2023strongsort,
  title={Strongsort: Make deepsort great again},
  author={Du, Yunhao and Zhao, Zhicheng and Song, Yang and Zhao, Yanyun and Su, Fei and Gong, Tao and Meng, Hongying},
  journal={IEEE Transactions on Multimedia},
  year={2023},
  publisher={IEEE}
}

@inproceedings{chen2024delving,
  title={Delving into the Trajectory Long-tail Distribution for Muti-object Tracking},
  author={Chen, Sijia and Yu, En and Li, Jinyang and Tao, Wenbing},
  booktitle={Proceedings of the IEEE/CVF Conference on Computer Vision and Pattern Recognition},
  pages={19341--19351},
  year={2024}
}

@inproceedings{meinhardt2022trackformer,
  title={Trackformer: Multi-object tracking with transformers},
  author={Meinhardt, Tim and Kirillov, Alexander and Leal-Taixe, Laura and Feichtenhofer, Christoph},
  booktitle={Proceedings of the IEEE/CVF conference on computer vision and pattern recognition},
  pages={8844--8854},
  year={2022}
}

@inproceedings{zeng2022motr,
  title={Motr: End-to-end multiple-object tracking with transformer},
  author={Zeng, Fangao and Dong, Bin and Zhang, Yuang and Wang, Tiancai and Zhang, Xiangyu and Wei, Yichen},
  booktitle={European Conference on Computer Vision},
  pages={659--675},
  year={2022},
  organization={Springer}
}

@article{yu2026rt,
  title={RT-RMOT: A Dataset and Framework for RGB-Thermal Referring Multi-Object Tracking},
  author={Yu, Yanqiu and Jin, Zhifan and Chen, Sijia and Chu, Tongfei and Yu, En and Liu, Liman and Tao, Wenbing},
  journal={arXiv preprint arXiv:2602.22033},
  year={2026}
}

@article{chamiti2025refergpt,
  title={ReferGPT: Towards Zero-Shot Referring Multi-Object Tracking},
  author={Chamiti, Tzoulio and Di Bella, Leandro and Munteanu, Adrian and Deligiannis, Nikos},
  journal={arXiv preprint arXiv:2504.09195},
  year={2025}
}

@article{liang2025cognitive,
  title={Cognitive Disentanglement for Referring Multi-Object Tracking},
  author={Liang, Shaofeng and Guan, Runwei and Lian, Wangwang and Liu, Daizong and Sun, Xiaolou and Wu, Dongming and Yue, Yutao and Ding, Weiping and Xiong, Hui},
  journal={Information Fusion},
  pages={103349},
  year={2025},
  publisher={Elsevier}
}

@article{li2025romot,
  title={ROMOT: Referring-expression-comprehension open-set multi-object tracking},
  author={Li, Wei and Li, Bowen and Wang, Jingqi and Meng, Weiliang and Zhang, Jiguang and Zhang, Xiaopeng},
  journal={The Visual Computer},
  volume={41},
  number={4},
  pages={2425--2437},
  year={2025},
  publisher={Springer}
}

@article{achiam2023gpt,
  title={Gpt-4 technical report},
  author={Achiam, Josh and Adler, Steven and Agarwal, Sandhini and Ahmad, Lama and Akkaya, Ilge and Aleman, Florencia Leoni and Almeida, Diogo and Altenschmidt, Janko and Altman, Sam and Anadkat, Shyamal and others},
  journal={arXiv preprint arXiv:2303.08774},
  year={2023}
}

@article{bai2025qwen2,
  title={Qwen2. 5-vl technical report},
  author={Bai, Shuai and Chen, Keqin and Liu, Xuejing and Wang, Jialin and Ge, Wenbin and Song, Sibo and Dang, Kai and Wang, Peng and Wang, Shijie and Tang, Jun and others},
  journal={arXiv preprint arXiv:2502.13923},
  year={2025}
}

@article{wang2025internvl3,
  title={Internvl3. 5: Advancing open-source multimodal models in versatility, reasoning, and efficiency},
  author={Wang, Weiyun and Gao, Zhangwei and Gu, Lixin and Pu, Hengjun and Cui, Long and Wei, Xingguang and Liu, Zhaoyang and Jing, Linglin and Ye, Shenglong and Shao, Jie and others},
  journal={arXiv preprint arXiv:2508.18265},
  year={2025}
}

@inproceedings{xiao2024florence,
  title={Florence-2: Advancing a unified representation for a variety of vision tasks},
  author={Xiao, Bin and Wu, Haiping and Xu, Weijian and Dai, Xiyang and Hu, Houdong and Lu, Yumao and Zeng, Michael and Liu, Ce and Yuan, Lu},
  booktitle={Proceedings of the IEEE/CVF Conference on Computer Vision and Pattern Recognition},
  pages={4818--4829},
  year={2024}
}

@article{zhang2025leader360v,
  title={Leader360V: The Large-scale, Real-world 360 Video Dataset for Multi-task Learning in Diverse Environment},
  author={Zhang, Weiming and Xiao, Dingwen and Dai, Aobotao and Liu, Yexin and Pan, Tianbo and Wen, Shiqi and Chen, Lei and Wang, Lin},
  journal={arXiv preprint arXiv:2506.14271},
  year={2025}
}

@inproceedings{he2021know,
  title={Know your surroundings: Panoramic multi-object tracking by multimodality collaboration},
  author={He, Yuhang and Yu, Wentao and Han, Jie and Wei, Xing and Hong, Xiaopeng and Gong, Yihong},
  booktitle={Proceedings of the IEEE/CVF Conference on Computer Vision and Pattern Recognition},
  pages={2969--2980},
  year={2021}
}

@inproceedings{luo2025omnidirectional,
  title={Omnidirectional Multi-Object Tracking},
  author={Luo, Kai and Shi, Hao and Wu, Sheng and Teng, Fei and Duan, Mengfei and Huang, Chang and Wang, Yuhang and Wang, Kaiwei and Yang, Kailun},
  booktitle={Proceedings of the Computer Vision and Pattern Recognition Conference},
  pages={21959--21969},
  year={2025}
}

@article{yu2023motrv3,
  title={Motrv3: Release-fetch supervision for end-to-end multi-object tracking},
  author={Yu, En and Wang, Tiancai and Li, Zhuoling and Zhang, Yuang and Zhang, Xiangyu and Tao, Wenbing},
  journal={arXiv preprint arXiv:2305.14298},
  year={2023}
}

@inproceedings{du2024ikun,
  title={ikun: Speak to trackers without retraining},
  author={Du, Yunhao and Lei, Cheng and Zhao, Zhicheng and Su, Fei},
  booktitle={Proceedings of the IEEE/CVF Conference on Computer Vision and Pattern Recognition},
  pages={19135--19144},
  year={2024}
}

@article{zhang2024bootstrapping,
  title={Bootstrapping Referring Multi-Object Tracking},
  author={Zhang, Yani and Wu, Dongming and Han, Wencheng and Dong, Xingping},
  journal={arXiv preprint arXiv:2406.05039},
  year={2024}
}

@article{meng2025motion,
  title={Motion-Perception Multi-Object Tracking (MPMOT): Enhancing Multi-Object Tracking Performance via Motion-Aware Data Association and Trajectory Connection},
  author={Meng, Weijun and Duan, Shuaipeng and Ma, Sugang and Hu, Bin},
  journal={Journal of Imaging},
  volume={11},
  number={5},
  pages={144},
  year={2025},
  publisher={MDPI}
}

@article{guan2025multi,
  title={Multi-object tracking review: retrospective and emerging trend},
  author={Guan, Zhiyu and Wang, Zhaofa and Zhang, Gan and Li, Luwei and Zhang, Miaomiao and Shi, Zhiping and Jiang, Na},
  journal={Artificial Intelligence Review},
  volume={58},
  number={8},
  pages={235},
  year={2025},
  publisher={Springer}
}

@inproceedings{chen2025cross,
  title={Cross-view referring multi-object tracking},
  author={Chen, Sijia and Yu, En and Tao, Wenbing},
  booktitle={Proceedings of the AAAI Conference on Artificial Intelligence},
  volume={39},
  number={2},
  pages={2204--2211},
  year={2025}
}

@article{lu2024deepseek,
  title={Deepseek-vl: towards real-world vision-language understanding},
  author={Lu, Haoyu and Liu, Wen and Zhang, Bo and Wang, Bingxuan and Dong, Kai and Liu, Bo and Sun, Jingxiang and Ren, Tongzheng and Li, Zhuoshu and Yang, Hao and others},
  journal={arXiv preprint arXiv:2403.05525},
  year={2024}
}

@article{kuhn1955hungarian,
  title={The Hungarian method for the assignment problem},
  author={Kuhn, Harold W},
  journal={Naval research logistics quarterly},
  volume={2},
  number={1-2},
  pages={83--97},
  year={1955},
  publisher={Wiley Online Library}
}

@inproceedings{radford2021learning,
  title={Learning transferable visual models from natural language supervision},
  author={Radford, Alec and Kim, Jong Wook and Hallacy, Chris and Ramesh, Aditya and Goh, Gabriel and Agarwal, Sandhini and Sastry, Girish and Askell, Amanda and Mishkin, Pamela and Clark, Jack and others},
  booktitle={International conference on machine learning},
  pages={8748--8763},
  year={2021},
  organization={PMLR}
}

@article{li2025ovtr,
  title={OVTR: End-to-End Open-Vocabulary Multiple Object Tracking with Transformer},
  author={Li, Jinyang and Yu, En and Chen, Sijia and Tao, Wenbing},
  journal={arXiv preprint arXiv:2503.10616},
  year={2025}
}

@misc{liu2024llavanext,
  title={Llavanext: Improved reasoning, ocr, and world knowledge},
  author={Liu, Haotian and Li, Chunyuan and Li, Yuheng and Li, Bo and Zhang, Yuanhan and Shen, Sheng and Lee, Yong Jae},
  year={2024}
}

@article{martin2019jrdb,
  title={Jrdb: A dataset and benchmark for visual perception for navigation in human environments},
  author={Mart{\'\i}n-Mart{\'\i}n, Roberto and Rezatofighi, Hamid and Shenoi, Abhijeet and Patel, Mihir and Gwak, J and Dass, Nathan and Federman, Alan and Goebel, Patrick and Savarese, Silvio},
  journal={arXiv preprint arXiv:1910.11792},
  year={2019}
}

@article{chen2026drmot,
  title={DRMOT: A Dataset and Framework for RGBD Referring Multi-Object Tracking},
  author={Chen, Sijia and Ma, Lijuan and Yu, Yanqiu and Yu, En and Liu, Liman and Tao, Wenbing},
  journal={arXiv preprint arXiv:2602.04692},
  year={2026}
}

@article{sun2024chattracker,
  title={Chattracker: Enhancing visual tracking performance via chatting with multimodal large language model},
  author={Sun, Yiming and Yu, Fan and Chen, Shaoxiang and Zhang, Yu and Huang, Junwei and Li, Yang and Li, Chenhui and Wang, Changbo},
  journal={Advances in Neural Information Processing Systems},
  volume={37},
  pages={39303--39324},
  year={2024}
}

@inproceedings{cao2023observation,
  title={Observation-centric sort: Rethinking sort for robust multi-object tracking},
  author={Cao, Jinkun and Pang, Jiangmiao and Weng, Xinshuo and Khirodkar, Rawal and Kitani, Kris},
  booktitle={Proceedings of the IEEE/CVF conference on computer vision and pattern recognition},
  pages={9686--9696},
  year={2023}
}
\end{document}